\documentclass[
]{ceurart}

\usepackage{listings}
\usepackage{soul}
\usepackage{xcolor}
\usepackage{xcolor}
\usepackage{pgfplots}
\usepackage{float}
\pgfplotsset{compat=1.18}
\usetikzlibrary{matrix, positioning}
\usepackage{todonotes}

\begin{document}

\copyrightyear{2025}
\copyrightclause{Copyright for this paper by its authors.
  Use permitted under Creative Commons License Attribution 4.0
  International (CC BY 4.0).}

\conference{}

\title{An analysis of vision-language models for fabric retrieval}

\author[]{Francesco Giuliari}[%
orcid=0000-0003-2311-9517,
email=fgiuliari@fbk.eu,
]
\cormark[1]
\fnmark[1]

\author[]{Asif Khan Pattan}[%
orcid=0009-0008-9871-5931,
email=apattan@fbk.eu,
]
\fnmark[1]

\author[]{Mohamed Lamine Mekhalfi}[%
orcid=0000-0002-4295-0974,
email=mmekhalfi@fbk.eu,
]
\fnmark[1]

\author[]{Fabio Poiesi}[%
orcid=0000-0002-9769-1279,
email=poiesi@fbk.eu,
]
\fnmark[1]

\address[]{Fondazione Bruno Kessler,
  Via Sommarive 18 - Povo, 38123 Trento, Italy}

\cortext[1]{Corresponding author.}
\fntext[1]{All authors contributed equally.}

\begin{abstract}
Effective cross-modal retrieval is essential for applications like information retrieval and recommendation systems, particularly in specialized domains such as manufacturing, where product information often consists of visual samples paired with a textual description. This paper investigates the use of Vision Language Models(VLMs) for zero-shot text-to-image retrieval on fabric samples. We address the lack of publicly available datasets by introducing an automated annotation pipeline that uses Multimodal Large Language Models (MLLMs) to generate two types of textual descriptions: freeform natural language and structured attribute-based descriptions. We produce these descriptions to evaluate retrieval performance across three Vision-Language Models: CLIP, LAION-CLIP, and Meta’s Perception Encoder. Our experiments demonstrate that structured, attribute-rich descriptions significantly enhance retrieval accuracy, particularly for visually complex fabric classes, with the Perception Encoder outperforming other models due to its robust feature alignment capabilities. However, zero-shot retrieval remains challenging in this fine-grained domain, underscoring the need for domain-adapted approaches. Our findings highlight the importance of combining technical textual descriptions with advanced VLMs to optimize cross-modal retrieval in industrial applications.
\end{abstract}


\maketitle

\section{Introduction} \label{Introduction}
The retrieval of relevant content from databases is a fundamental task crucial for applications like information retrieval and recommendation systems. Effective retrieval enables efficient access to the vast amounts of daily generated data, allowing users to quickly find matching information or items \cite{zhu2023large}.

Recent advancements in deep learning, particularly the development of aligned language and visual representations through contrastive pretraining~\cite{CLIP}, have significantly improved cross-modal matching between images and textual descriptions~\cite{beltran2021deep}. This progress enables efficient retrieval of visual information using text queries and vice versa, opening new possibilities for applications in various domains. In the manufacturing industry, where vast amounts of data are available in the form of production sample images, such as those featured in online stores, this capability allows for seamless retrieval of visual content based on textual descriptions, streamlining search and analysis processes.

Building on our work within the PNRR-funded ``Intrecci Digitali'' project, this paper focuses on \mbox{zero-shot} \mbox{text-to-image} retrieval, investigating how manipulating textual queries can enhance the retrieval accuracy of fabric samples.
To conduct our study on text-to-image retrieval in the fabric domain, we identified a lack of suitable publicly available datasets. To address this, we utilized an existing fabric image dataset and generated corresponding textual prompts to enable our experiments. Specifically, we explore the use of MLLMs for automatically generating detailed image descriptions, allowing for a more robust evaluation of retrieval performance. We consider multiple types of textual prompts, including freeform and template-based descriptions incorporating fabric-specific attributes.

Our study highlights that to achieve the best retrieval accuracy, it is necessary to use both a technical attribute-based description and a powerful model that can effectively use such a description.

\begin{figure}[t]
\centering
\includegraphics[width=\linewidth]{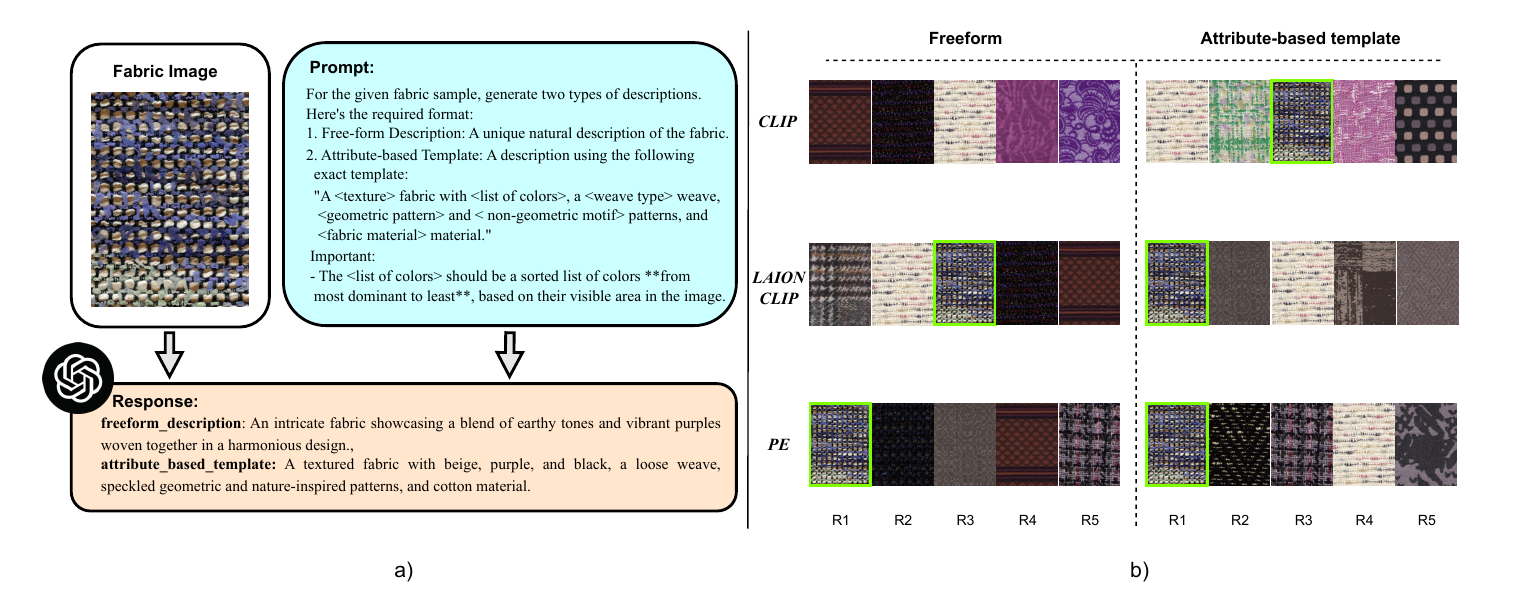}
\label{fig:prompts_result}
\caption{Visualization of our descriptions generation and retrieval process. (a) Shows the input fabric image, the prompt given to ChatGPT, and the resulting generated description. (b) Displays the Top-5 image retrieval results using two types of generated descriptions across three models: CLIP, LAION-CLIP, and the Perception Encoder (PE). R1, R2, etc., indicate the ranked order of retrieved images based on similarity scores.}
\end{figure}

\section{Related Works}

\noindent\textbf{Text-to-Image Retrieval}
The development of VLMs like CLIP \cite{CLIP} has enabled effective cross-modal retrieval by learning to associate images and text within a shared embedding space. Trained on large-scale image-text datasets, these models align visual and textual information simultaneously, supporting tasks such as text-to-image retrieval. Earlier works \cite{wang2024fabric, xiang2021efficient} in fabric image retrieval have primarily focused on image-to-image retrieval and are often limited to one or two specific attributes (e.g., texture, color). These approaches struggle with complex fabrics like printed textiles, where texture alone is insufficient to capture the full visual diversity.  Recent advances in multimodal understanding-such as the pipeline proposed in \cite{liu2024democratizing}-demonstrate the effectiveness of combining MLLMs with VLMs for fine-grained visual reasoning. This integration enhances semantic alignment across modalities, which is crucial for cross-modal retrieval tasks requiring nuanced interpretation of visual attributes. Only a few recent studies, such as \cite{suzuki2023text}, have explored text-to-image fabric retrieval, proposing a training-based framework that leverages VLMs (e.g., CLIP \cite{CLIP}). In contrast, this work analyzes the zero-shot performance of VLMs on domain-specific cross-modal retrieval tasks such as fabric image retrieval, considering multiple visual attributes. Related analyses, such as \cite{sultan2023exploring}, have shown that including expressive terms reflecting the 'tone' of a scene can enhance the alignment between text and image representations in natural images, suggesting the importance of descriptive richness in query formulation.

\section{Analysis of cross-modal fabric retrieval} \label{Analysis of cross-modal fabric retrieval}

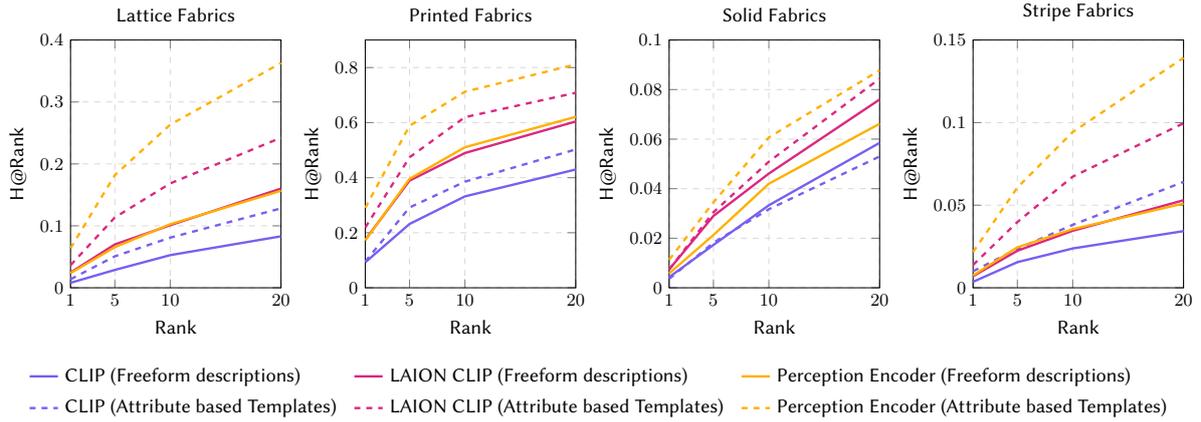
\begin{figure}[t]
\centering
{
\definecolor{ModelAColor}{HTML}{785EF0} %
\definecolor{ModelBColor}{HTML}{DC267F} %
\definecolor{ModelCColor}{HTML}{FFB000} %
\newcommand{\ModelA}{Model Alpha}
\newcommand{\ModelB}{Model Beta}
\newcommand{\ModelC}{Model Gamma}
\newcommand{\SettingOne}{Setting 1}
\newcommand{\SettingTwo}{Setting 2}
\newcommand{\SettingThree}{Setting 3}
\tikzset{
    setting one/.style={solid,very thick},
    setting two/.style={dashed, thick},
    setting three/.style={dashed,very thick},
}
\resizebox{\linewidth}{!}{%
\begin{tikzpicture}
\matrix[draw=none,row sep=0cm, column sep=0.01cm,ampersand replacement=\&] {
        \begin{axis}[
        title=Lattice Fabrics,
        width=\textwidth/3,
        height=6cm,
        xlabel={Rank},
        ylabel={H@Rank},
        xmin=1, xmax=20,
        ymin=0, ymax=0.4,
        xtick={1,5,10,20},
        label style={font=\small},
        tick label style={font=\footnotesize},
        grid=major,
        grid style={dashed, gray!30},
        yticklabel style={
        /pgf/number format/fixed,
        /pgf/number format/precision=3
}
    ]
    \addplot[ModelAColor, setting one] coordinates {(1,0.008)(5,0.0288)(10,0.0527)(20, 0.0831)};
    \addplot[ModelAColor, setting three] coordinates {(1,0.0137)(5,0.0508)(10,0.0809)(20, 0.1282)};
    \addplot[ModelBColor, setting one] coordinates {(1,0.0246)(5,0.07)(10,0.101)(20, 0.1602)};
    \addplot[ModelBColor, setting three] coordinates {(1,0.0364)(5,0.1135)(10,0.1685)(20, 0.2423)};
    \addplot[ModelCColor, setting one] coordinates {(1,0.0237)(5,0.0655)(10,0.1026)(20, 0.1566)};
    \addplot[ModelCColor, setting three] coordinates {(1,0.0639)(5,0.1819)(10,0.2634)(20, 0.3629)};
    \end{axis} \& \begin{axis}[
        title=Printed Fabrics,
        width=\textwidth/3,
        height=6cm,
        xlabel={Rank},
        ylabel={H@Rank},
        xmin=1, xmax=20,
        ymin=0, ymax=0.9,
        xtick={1,5,10,20},
        label style={font=\small},
        tick label style={font=\footnotesize},
        grid=major,
        grid style={dashed, gray!30},
        yticklabel style={
        /pgf/number format/fixed,
        /pgf/number format/precision=3
}
    ]
    \addplot[ModelAColor, setting one] coordinates {(1,0.0924)(5,0.2318)(10,0.332)(20, 0.4297)};
    \addplot[ModelAColor, setting three] coordinates {(1,0.0951)(5,0.2917)(10,0.3854)(20, 0.5026)};
    \addplot[ModelBColor, setting one] coordinates {(1,0.1745)(5,0.3893)(10,0.4896)(20, 0.6042)};
    \addplot[ModelBColor, setting three] coordinates {(1,0.2188)(5,0.474)(10,0.6198)(20, 0.7083)};
    \addplot[ModelCColor, setting one] coordinates {(1,0.1719)(5,0.3958)(10,0.5104)(20, 0.6211)};
    \addplot[ModelCColor, setting three] coordinates {(1,0.2904)(5,0.5885)(10,0.7122)(20, 0.8112)};
    \end{axis}
    \& \begin{axis}[
        title=Solid Fabrics,
        width=\textwidth/3,
        height=6cm,
        xlabel={Rank},
        ylabel={H@Rank},
        xmin=1, xmax=20,
        ymin=0, ymax=0.1,
        xtick={1,5,10,20},
        label style={font=\small},
        tick label style={font=\footnotesize},
        grid=major,
        grid style={dashed, gray!30},
        yticklabel style={
        /pgf/number format/fixed,
        /pgf/number format/precision=3
}
    ]
    \addplot[ModelAColor, setting one] coordinates {(1,0.0043)(5,0.0173)(10,0.0333)(20, 0.0585)};
    \addplot[ModelAColor, setting three] coordinates {(1,0.0036)(5,0.0182)(10,0.0317)(20, 0.053)};
    \addplot[ModelBColor, setting one] coordinates {(1,0.0074)(5,0.029)(10,0.0461)(20, 0.076)};
    \addplot[ModelBColor, setting three] coordinates {(1,0.0074)(5,0.0305)(10,0.0509)(20, 0.0844)};
    \addplot[ModelCColor, setting one] coordinates {(1,0.0058)(5,0.0213)(10,0.042)(20, 0.0662)};
    \addplot[ModelCColor, setting three] coordinates {(1,0.0113)(5,0.0345)(10,0.0607)(20, 0.0878)};
    \end{axis}
    \& \begin{axis}[
        title=Stripe Fabrics,
        width=\textwidth/3,
        height=6cm,
        xlabel={Rank},
        ylabel={H@Rank},
        xmin=1, xmax=20,
        ymin=0, ymax=0.15,
        xtick={1,5,10,20},
        label style={font=\small},
        tick label style={font=\footnotesize},
        grid=major,
        grid style={dashed, gray!30},
        yticklabel style={
        /pgf/number format/fixed,
        /pgf/number format/precision=3
}
    ]
    \addplot[ModelAColor, setting one] coordinates {(1,0.0036)(5,0.0155)(10,0.0238)(20, 0.0343)};
    \addplot[ModelAColor, setting three] coordinates {(1,0.01)(5,0.0236)(10,0.0381)(20, 0.0641)};
    \addplot[ModelBColor, setting one] coordinates {(1,0.007)(5,0.0224)(10,0.0345)(20, 0.053)};
    \addplot[ModelBColor, setting three] coordinates {(1,0.0138)(5,0.0398)(10,0.0673)(20, 0.0996)};
    \addplot[ModelCColor, setting one] coordinates {(1,0.0075)(5,0.0245)(10,0.0355)(20, 0.051)};
    \addplot[ModelCColor, setting three] coordinates {(1,0.0216)(5,0.0607)(10,0.0943)(20, 0.1392)};
    \end{axis}\\
    };
    \node[below=0cm of current bounding box.south] {
    \begin{tikzpicture}
    \matrix [draw=none, row sep=0.01pt, column sep=5pt,ampersand replacement=\&] {
        \draw[ModelAColor, setting one] (-0.5,0) -- (0,0); \node[right] {CLIP (Freeform descriptions)}; \&
        \draw[ModelBColor, setting one] (-0.5,0) -- (0,0); \node[right] {LAION CLIP (Freeform descriptions)}; \&
        \draw[ModelCColor, setting one] (-0.5,0) -- (0,0); \node[right] {Perception Encoder (Freeform descriptions)}; \\
        \draw[ModelAColor, setting three] (-0.5,0) -- (0,0); \node[right] {CLIP (Attribute based Templates)}; \&
        \draw[ModelBColor, setting three] (-0.5,0) -- (0,0); \node[right] {LAION CLIP (Attribute based Templates)}; \&
        \draw[ModelCColor, setting three] (-0.5,0) -- (0,0);\node[right] {Perception Encoder (Attribute based Templates)}; \\
    };
    \end{tikzpicture}
    };
\end{tikzpicture}%
}
}%
\caption{Retrieval Hit-rate analysis on the four fabric classes of FID. We compare three models: CLIP, LAION-CLIP, and Perception Encoder, using both Freeform descriptions and an Attribute-based template description.}\label{fig:result_plots}
\end{figure}

In cross-modal text-to-image retrieval, the goal is to identify an image in a database that corresponds to a given text description. Quantitatively evaluating this task requires high-quality data, as each image must be paired with a description that uniquely distinguishes it from all others. This challenge is particularly difficult for fabric images due to several factors: the lack of publicly available annotated datasets, the underrepresentation of fabric images in large-scale VLMs training data, and the inherent visual similarity among fabric samples, which makes them difficult to describe in a distinctive manner.
To address the missing annotations in fabric image datasets, we introduce an automatic annotation pipeline that generates two distinct description types using LLMs: freeform natural language descriptions that describe the fabric holistically, and structured technical descriptions where predefined attribute templates are populated.
In this study, we evaluate three VLMs using both description paradigms on fabric imagery, determining which combination of textual representation and embedding model proves most effective for this specific retrieval task.

\noindent\textbf{Dataset.} We use the Fabric-Image-Data (FID)\footnote{\url{https://github.com/rhrobot/Fabric-Image-Data}}\cite{liu2024improved}, which comprises 12,181 wool fabric images categorized into four splits: lattice (3,128 images), pattern (768 images), solid (4,169 images), and stripe (4,116 images). Each image has a resolution of $420 \times 570$ pixels. While the original dataset is designed for fabric image classification, we adapt it for the task of text-to-image retrieval. To support this task, we generate two types of textual descriptions for each image. The detailed description generation process and the prompts used are described in the following section.

\noindent\textbf{Description Generation.} \label{desc_gen} Descriptions are generated using ChatGPT-4o-mini based on the input prompt shown in Figure \ref{fig:prompts_result}a. For the 'Attribute-based template' description, the attributes are selected from ChatGPT-4o-mini's response to a separate, image-independent prompt: \textit{"List the top 5 attributes to distinguish a fabric image."} On average, the model generates each response in approximately 2.6 seconds.

\noindent\textbf{Experimental Setup.} We evaluate the retrieval performance of the models using both types of descriptions. For each model, we first pre-compute and store the image embeddings for all images in a given split. Then, for each textual description, we extract its text embedding using the model’s text encoder. We compute the cosine similarity between the description’s embedding and every image embedding in the split, ranking the results in descending order based on similarity.
We report the results in terms of Hit-Rate @ Rank $K$ ($H@K$), where the $H@K$ for each query is $1$ if the image corresponding to the description is among the first $K$ images and 0 otherwise. The final score is computed by averaging $H@K$ across all descriptions in the split.  We report the scores with ranks: 1,5,10, and 20.

\noindent\textbf{Compared Vision-Language Models.} We compare three text-image embedding models: the CLIP\footnote{\url{https://huggingface.co/openai/clip-vit-large-patch14}}\cite{CLIP} model from OpenAI; LAION-CLIP\footnote{\url{https://huggingface.co/laion/CLIP-ViT-H-14-laion2B-s32B-b79K}}, which is a version of the CLIP model trained on LAION2B data\cite{schuhmann2022laionb}; and Perception Encoder\footnote{\url{https://huggingface.co/facebook/PE-Core-L14-336}}\cite{bolya2025perception}, a recent model from Meta for image and text alignment.

\noindent\textbf{Results Discussion.} We evaluate the retrieval performance of three pre-trained models (CLIP, LAION-CLIP, and Perception Encoder) using two types of automatically generated descriptions: free-form and attribute-based template. Attribute-based descriptions were chosen for their ability to encode structured, fine-grained visual cues-such as color, weave, and pattern-that are critical for distinguishing highly similar fabric images. As shown in Figure \ref{fig:result_plots}, these structured descriptions lead to consistently higher retrieval accuracy across all models, with the greatest improvements observed in visually complex fabric classes like "lattice" and "printed." Among the models, the Perception Encoder delivers the strongest performance, benefiting significantly from attribute-based inputs. This suggests that its strong performance is driven by both a large training corpus and the use of intermediate feature representations, which enhance visual-text alignment through richer embeddings.

\section{Conclusions}

In this study, we report our findings on using pretrained Vision Language Models for \mbox{Text-to-Image} \mbox{cross-modal retrieval} on data from the fabric industry. Given the lack of manually annotated data in this domain, we propose a framework for the automatic labeling of fabric samples via the use of Chat-GPT to evaluate the performance of Text-to-Image retrieval systems.

Our findings highlight two critical factors for optimizing retrieval accuracy: First, retrieval accuracy improves substantially when fabric descriptions are technical and structured, incorporating details such as color, weave type, and patterns, rather than freeform, as evidenced by consistent gains across all tested models. Second, model choice plays a decisive role: while CLIP, the de facto standard academic VLM, performs poorly on fabric data even with augmented training (e.g., Laion-CLIP), the Perception encoder proves more adept at extracting relevant features from structured textual inputs. 
Nonetheless, absolute retrieval accuracy still is limited. Even with the most advanced model and descriptive input, zero-shot retrieval on fine-grained, domain-specific data such as fabrics remains a challenging task. In future works, we will continue to explore ways to improve performance in this particular domain.

\begin{acknowledgments}
This work was carried out as part of the Intrecci Digitali project, developed within the research program PE00000004 `Made in Italy Circolare e Sostenibile – MICS', funded by the PNRR – Missione 4, Componente 2, Linea 1.3 – CUP D43C22003120001 \end{acknowledgments}

\section*{Declaration on Generative AI}
During the preparation of this work, the authors used ChatGPT and Deepseek in order to: Grammar and spelling check, Improve writing style, Paraphrase and reword. After using these tools, the authors reviewed and edited the content as needed and take full responsibility for the publication’s content.

\bibliography{References}

\end{document}